\documentclass[a4]{esannV2}
\usepackage[dvips]{graphicx}
\usepackage[latin1]{inputenc}
\usepackage{amssymb,amsmath,array}
\usepackage{caption}
\usepackage{subcaption}
\usepackage{color}
\usepackage{xspace}
\usepackage{cite}

\newcommand{\teacher}{\texttt{teacher}\xspace}
\newcommand{\student}{\texttt{student}\xspace}

\begin{document}
%style file for ESANN manuscripts
\title{Knowledge Distillation for Anomaly Detection}

\author{
Adrian Alan Pol$^{2*}$ %
Ekaterina Govorkova$^{3*}$ %
Sonja Gr\"{o}nroos$^{1*}$ %
Nadezda Chernyavskaya$^4$\\%
Philip Harris$^3$ 
Maurizio Pierini$^4$ 
Isobel Ojalvo$^2$ 
Peter Elmer$^2$ % 
$^*${\footnotesize \textit{equal contribution}}
% % Optional short acknowledgment: remove next line if non-needed
% \thanks{This is an optional funding source acknowledgement.}
%
% DO NOT MODIFY THE FOLLOWING '\vspace' ARGUMENT
\vspace{.3cm}\\
$^1$ University of Helsinki, Finland \\
$^2$ Princeton University, USA \\
$^3$ Massachusetts Inst. of Technology, USA \\
$^4$ European Organization for Nuclear Research (CERN), Switzerland \\
}

\maketitle

\begin{abstract}
Unsupervised deep learning techniques are widely used to identify anomalous behaviour. The performance of such methods is a product of the amount of training data and the model size. However, the size is often a limiting factor for the deployment on resource-constrained devices. We present a novel procedure based on knowledge distillation for compressing an unsupervised anomaly detection model into a supervised deployable one and we suggest a set of techniques to improve the detection sensitivity. Compressed models perform comparably to their larger counterparts while significantly reducing the size and memory footprint.
\end{abstract}

\section{Introduction}
\label{introduction}
The unsupervised techniques use unlabeled training data to help uncover out-of-distribution samples. Unsupervised anomaly and novelty detection are essential in various domains, e.g. in particle physics they were used to detect abnormal behaviour of detector components or to search for new physics phenomena
~\cite{pol2018detector, xiao2020large, Govorkova_2022}.

Recently, the advances in anomaly detection were dominated by deep learning techniques, especially autoencoders which can overcome large dimensionality and non-linear relationships in input data without labels. These new methods can require a complex model to achieve high performance, which can prevent their deployment in resource-constrained environments, such as edge devices or embedded systems. Whether a candidate model is suitable for deployment in such cases depends on meeting the production demands, which include performance and inference requirements, e.g. area, power or latency. We are especially interested in the high energy physics applications where $100$~ns latency is desired. However, we first validate our approach on common machine learning datasets.

We propose a novel method for compressing an unsupervised anomaly detector into a small, deployable model. Our strategy leverages knowledge distillation~(KD)~\cite{bucilua2006model, hinton2015distilling}, a method in which a large \teacher model transfers its knowledge to a smaller \student model, without sacrificing performance measured by top metrics such as the area under the receiver operating characteristic (ROC) curve (ROC-AUC). We expand the traditional KD setup by turning an unsupervised problem into a supervised one. By doing so, we reduce the complexity of the task that the \student needs to solve, i.e. by dropping the dimensionality of the output to a single number. This reduction leads to the \student directly learning the anomaly metric rather than a typical autoencoder approach. Finally, to address concerns of~\cite{stanton2021does} we suggest extensions to improve generalization.

\section{Related Work and Methodology}
\label{sec:methodology}

The intermediate feature maps can be used to train the \student models. This KD setup was previously used for anomaly detection tasks. While~\cite{salehi2021multiresolution} followed this idea to train a shallower model, \cite{liu2022unsupervised} utilized this setup to mitigate the impact of anomalies in the training set and \cite{xiao2021unsupervised} used the low-dimensional embedding of the \teacher to guide the training of an ensemble of students. In the following, we propose to skip the embedding learning and directly regress the anomaly score reported by the teacher. These two approaches can be complementary but we leave empirical verification of such combination for future work.

\begin{figure}[htb]
\centering
\includegraphics[width=0.8\textwidth]{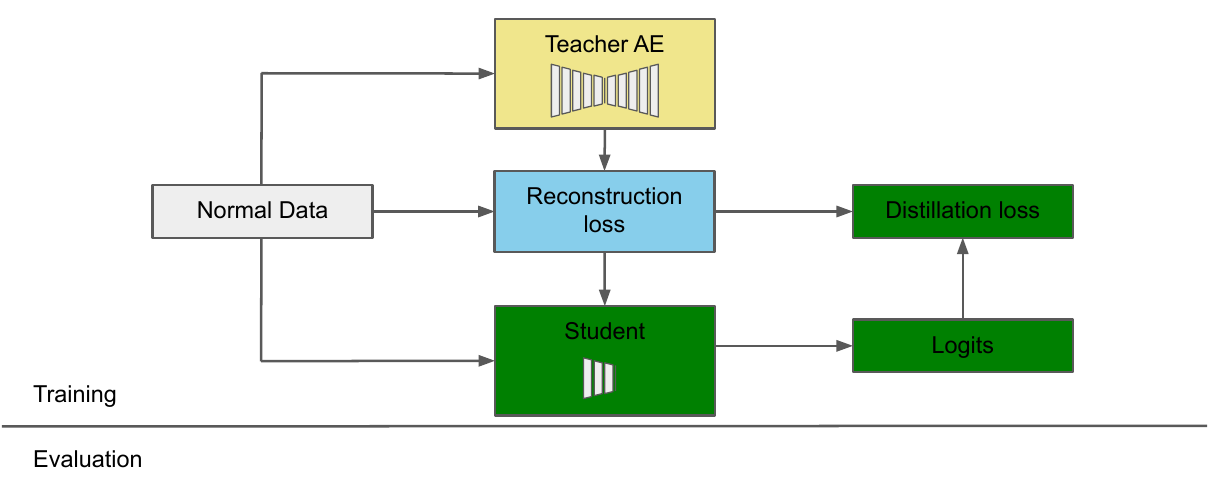} \\
\includegraphics[width=0.75\textwidth]{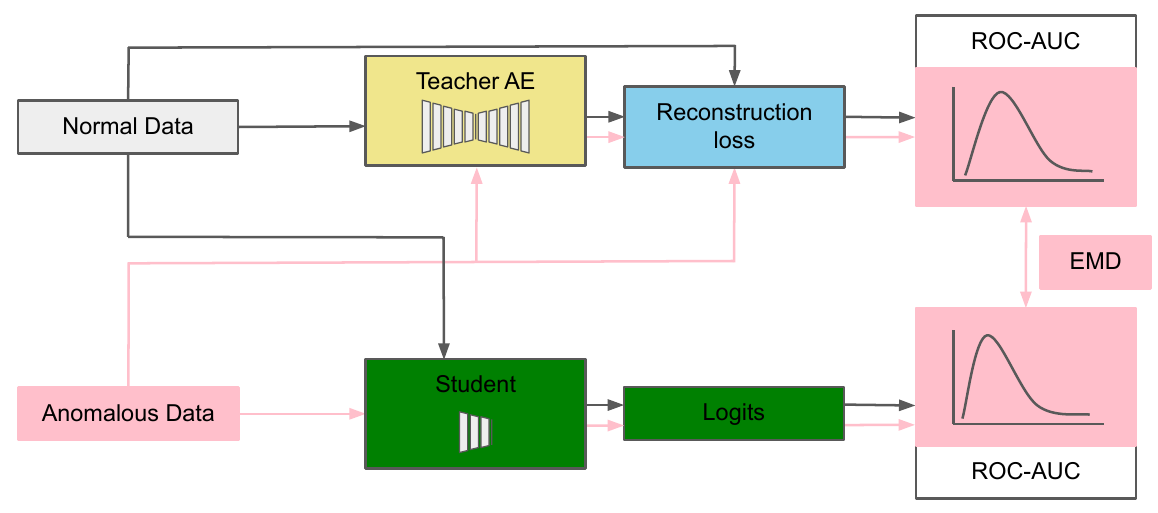} 
\caption{The schematic view of the baseline training and evaluation procedure.}
\label{fig:method}
\end{figure}

Consider an unlabelled dataset $D = \{(x)_{i}\}_{1}^{N}$, free from outliers. We search for a parameterized function $f_\theta$ that produces high scores $\mathcal{S}$ for outliers and low for inliers. We approximate $f_\theta$ using the loss $\mathcal{L}$ of an autoencoder with encoder $\mathcal{E}$ and decoder $\mathcal{D}$. We calculate the anomaly score as $\mathcal{S}_i =  f_\theta(x_i) = \mathcal{L}(x_i,\mathcal{D}(\mathcal{E}(x_i))$. The training and evaluation schemes are shown in Fig.~\ref{fig:method}.
% In order to alleviate the size concerns we develop a teacher model as an intermediate step towards a deployable model. 
% We use it for unsupervised anomaly detection as described above. After training the model only on normal samples, we evaluate the full test set (both normal and anomalous samples) and compute the anomaly scores, i.e. the reconstruction loss.
First, we train a \teacher model defined as an autoencoder, commonly used for unsupervised anomaly detection on image inputs. 
% After training the model on normal samples, we apply it to both normal and anomalous samples and compute the loss $s$ between the input and the network outputs. 
Next, we introduce the \student model $g$ and train it directly to learn the anomaly score: $g(x_i)\approx \mathcal{S}_i$. The \student architecture differs from the \teacher's as it is both simpler and not an autoencoder.
%The \student, $g$, learns the anomaly score $s$ reported by the teacher directly  without the need to reconstruct the input (no decoder): $g(X) \approx s$.
This allows the \student to leverage the knowledge of the \teacher more efficiently to reduce the computational overhead and complexity of the model. 
By not requiring to reconstruct the input, we can focus its learning on the most important aspects of the data, enhancing its ability to detect anomalies.

\begin{table}[ht]
\footnotesize
\centering
\begin{tabular}{|l|c|c|c|c|c|c|c|c|}
\hline
 & {\scriptsize Teacher} & {\scriptsize S$_1$} & {\scriptsize  S$_2$} & {\scriptsize S$_3$} & {\scriptsize  S$_4$} & {\scriptsize  S$_5$} & {\scriptsize S$_6$} & {\scriptsize S$_7$} \\ \hline
{\scriptsize Params} & 19,360 & 7,180 & 2190 &1060	 &409 & 225 &133 &77 \\
{\scriptsize FLOPS} & 18.91M & 6.37M &	2.25M&432k & 131k &114k &58k &53k \\ \hline
\end{tabular}
\caption{Comparison of a number of parameters and FLOPS for \teacher and \student models (numbered as S$_n$, where $n \in \{1, 2, ..., 7\}$).}
\label{tab:archs}
\end{table}

\subsection{Models Architecture and Training}
\label{subsec:architecture}

The architecture of the \teacher consists of five convolutional layers with average pooling layers in both the encoder and decoder, a representation vector size of 20 and two fully connected layers in between.

To find out the size impact on \student performance, we experiment with seven different architectures as outlined in Table~\ref{tab:archs}. 
All \student models are significantly smaller than the \teacher, with the largest size of an encoder-only part of the \teacher. Results of ~\cite{abnar2020transferring} suggested that inductive biases could be transferred in the context of KD. However, we opted for experimenting only with convolutional-based students. We refer to these students as $S_n$, where $n \in [1, 7]$. S$_1$ and S$_2$ consist of five, S$_3$ of three and S$_4$, S$_5$, S$_6$, S$_7$ of two convolutional layers followed by one dense layer. 

We train our \teacher models on two widely used datasets in experimental settings, MNIST and Fashion-MNIST, using an unsupervised learning approach: the images of one class were treated as the normal training dataset at a time, i.e. 6000 training samples.
Twenty teacher models are trained for both datasets: one for each class of normal digits. The models are optimized using the mean squared error (MSE) loss function and trained using the $Adam$ optimizer with an initial learning rate of 10$^{-3}$.
We apply log transformation to \teacher loss to facilitate \student learning. 
The scores obtained for normal digits peak at zero with a long tail, which is why the log transformation transforms skewed targets to approximately conform to normality. 
An initial learning rate of 10$^{-3}$ with $Adam$ is used for all the \student models. We train all the models using mean absolute error (MAE) for $300$ epochs with a batch size of $100$.

% The task of anomaly detection with AEs requires to have a balance between over- and under-parameterized models such that the AE is not able to generalize to anomalous events, but is still able to learn dependencies and correlations in the normal dataset effectively.
Since the \student models are trained to directly regress the loss of the \teacher, we have to check their ability to correctly learn the scores of out-of-distribution samples, i.e. anomalies. In the baseline setting, the \student is only exposed to a modicum of high-loss examples during the training. Thus, during evaluation in addition to checking the detection performance, we check the ability of the \student to correctly reproduce the scores on anomalous samples as well. We quantify this ability by comparing the distance between the loss distribution of the \teacher and \student using Wasserstein distance~\cite{emd}, also referred to as Earth Mover's Distance~(EMD). 

\subsection{Co-learning of \teacher and \student}
\label{subsec:online_kd}
We explore simultaneous learning of models and refer to it as co-learning. The \teacher model's outputs are used as soft targets for the \student model.
We minimize a joint of \teacher reconstruction loss and \student distillation loss. 

\subsection{Outlier Exposure}
\label{subsec:outliers}
To expose the \student to samples that have different data distribution than the inlier class, we use the events of the MNIST dataset as outliers for the teacher trained on Fashion-MNIST and the other way around. 
This is the situation that is possible in real-life experiments when anomalous examples are not available but some other unrelated dataset is at hand. 
After training the \teacher on a normal digit, we train the \student on the blend of the normal digit and some additional samples from this different, unrelated dataset. This strategy is inspired by~\cite{kulkarni2017knowledge} and we refer to it as {\em outlier exposure}. Outlier exposure is widely used in the anomaly detection field both by scholars and practitioners.

\subsection{Denoising as Outlier Exposure}
\label{subsec:denoising}
Another way of outlier exposure we experiment on is noise addition.
Denoising is often used to help generalization and it was previously used for KD~\cite{sau2016deep}.
We apply noise only to the student training set only: $x + \epsilon \times \mathcal{N}(0, 1)$.
By being exposed to noisy data, we hope the students can better match the outlier distribution. 
Unlike the method proposed above, the outlier exposure with noised samples does not require any additional dataset. 
We use a noise factor of $\epsilon = 0.1$. 

\section{Experimental Results}
\label{sec:results}

We evaluate the performance of compressed models using ROC-AUC and EMD as evaluation metrics on a balanced $2000$-sample test set, i.e. containing $1000$ anomalies.
%Each test sample loss corresponds to the anomaly score which we use to calculate the true positive and false positive rate.
We compare the performance of \student models with that of the \teacher and analyze the trade-off between the model size and performance. The problem of anomaly detection and its compression is very problem-specific and architecture-specific. That is why we choose to repeat our experiments on multiple architectures and use each out of $10$ available settings, i.e. different {\em inlier} classes. This yields over $500$ trained models and corresponding evaluations. To discover trends we summarize our results and findings below.

We first train the baseline, i.e. an offline learning setup where teacher training is done before student learning. We then attempt co-learning and the results on average improve, both in terms of ROC-AUC and EMD. Therefore for the subsequent studies, we also use the co-learning setup. Next, we attempt to perform an outlier exposure outlined in Section~\ref{subsec:outliers}. This improves the results further, especially the distribution distance. This is expected since \student models can learn from out-of-distribution datasets. Finally, we find out that noise injection, see Section~\ref{subsec:denoising}, is a double edge sword. Despite the reduction of the outlier distribution distance, this setting simultaneously increases the inlier distribution distance and results in overall decreased sensitivity to anomalies.

\begin{figure}[h]
  \centering
  \includegraphics[width=.49\linewidth]{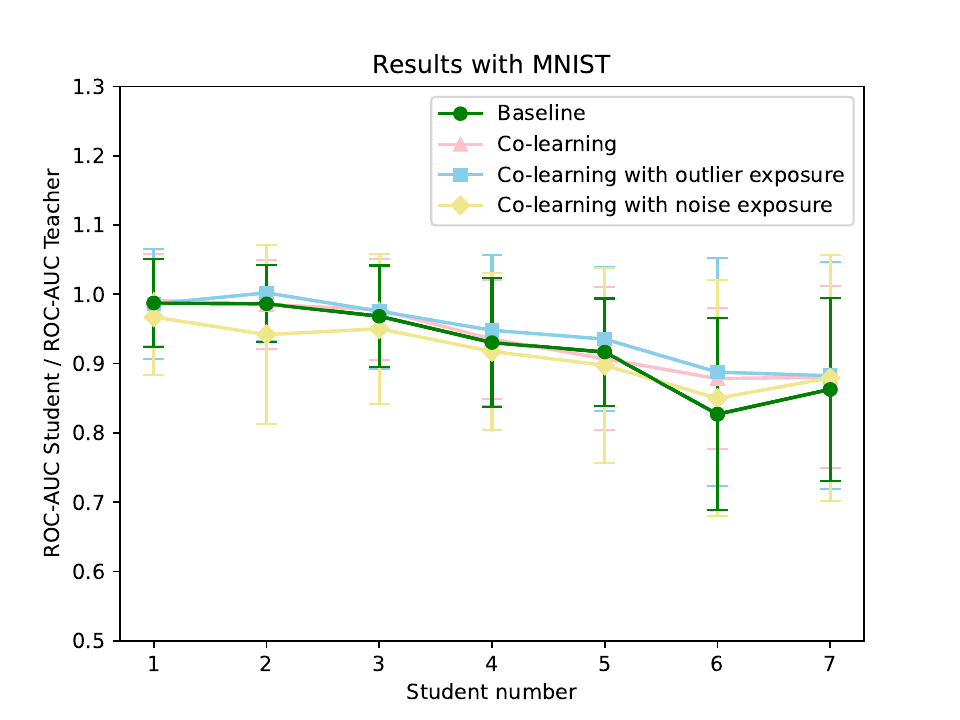}
  \includegraphics[width=.49\linewidth]{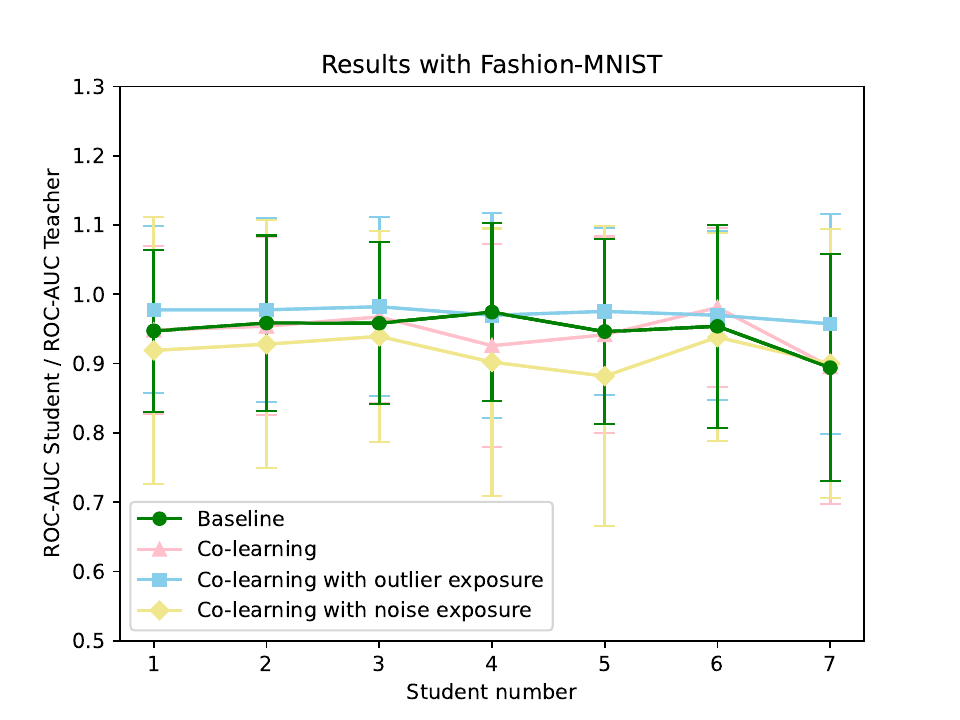} \\
  \includegraphics[width=.49\linewidth]{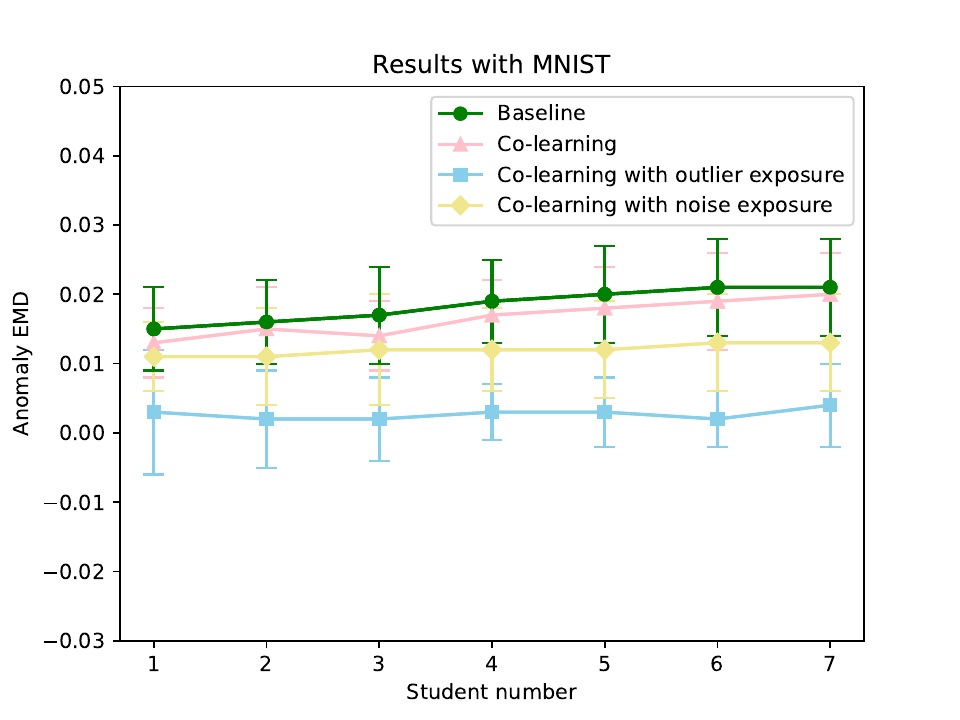}
  \includegraphics[width=.49\linewidth]{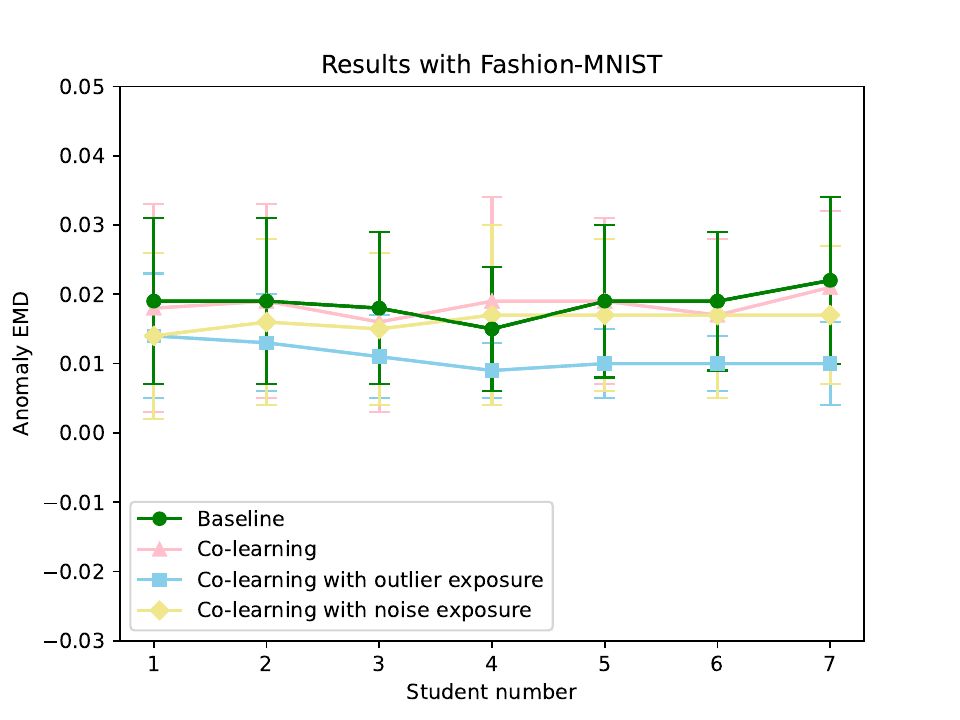}
\caption{On the top are ROC-AUC ratios and on the bottom are distances between the \student and the \teacher model obtained on the MNIST (left) and Fashion-MNIST (right) datasets. The results are compared between four different approaches: naive KD with loss regression~(green), co-learning of the \teacher and \student~(pink), co-learning with additional outlier exposure~(blue) and co-learning with noise outlier exposure~(yellow).}
\label{fig:metric}
\end{figure}

The ratio of ROC-AUC and EMD is presented for all \student models in Fig.~\ref{fig:metric}. The results are averaged between all 10 experiments and we observe a general trend: the performance of the \student models increases with the number of parameters and the outlier EMD decreases. Besides higher capacity, the capacity gap decreases which was shown to degrade knowledge transfer~\cite{mirzadeh2020improved}. This is seen in the results obtained with the MNIST dataset. However, results obtained on the Fashion-MNIST are too noisy to be interpreted. Another observation is that the co-learning together with outlier exposure produces on average the best results, both in terms of ROC-AUC and EMD.

The results demonstrate the effectiveness of the proposed approach for compressing unsupervised anomaly detectors using KD. Compressed models achieved significant reductions in model size while maintaining high detection sensitivity. 

\section{Conclusions}
\label{sec:conclusions}
In this paper, we proposed a novel technique for compressing an unsupervised anomaly detector into a small, deployable model using KD. Our approach leverages the knowledge of a larger \teacher model to train a smaller \student model, effectively reducing the model size and complexity without sacrificing performance. We validated our approach on two widely used datasets, MNIST and Fashion-MNIST, and demonstrated that \student models achieved comparable performance to their larger counterparts while significantly reducing the model size and memory footprint. We also discussed the impact of outlier exposure.

Although we identified several general trends in our results, we must stress that they will always be dataset and architecture dependent. In future work, we plan to investigate the applicability of our approach to application-specific use cases. We also plan to evaluate the performance of our compressed models on other evaluation metrics, such as computational efficiency and energy consumption, to further validate the practicality of our method. 

% \section*{Acknowledgements}
% This work is supported by the European Research Council (ERC) under the European Union's Horizon 2020 research and innovation program (Grant Agreement No. 772369) and the ERC-POC programme (grant No. 996696).

% % ****************************************************************************
% % BIBLIOGRAPHY AREA
% % ****************************************************************************

\begin{footnotesize}
\bibliographystyle{unsrt}
\bibliography{biblio}

\begin{thebibliography}{10}

\bibitem{pol2018detector}
AA~Pol, G~Cerminara, C~Germain, M~Pierini, and A~Seth.
\newblock {Detector monitoring with artificial neural networks at the CMS
  experiment at the CERN Large Hadron Collider}, 2018.

\bibitem{xiao2020large}
H~Xiao, I~Grama, and Q~Liu.
\newblock Large deviation expansions for the coefficients of random walks on
  the general linear group, 2020.

\bibitem{Govorkova_2022}
E~Govorkova, E~Puljak, T~Aarrestad, T~James, V~Loncar, M~Pierini, AA~Pol,
  N~Ghielmetti, M~Graczyk, S~Summers, J~Ngadiuba, TQ~Nguyen, J~Duarte, and
  Z~Wu.
\newblock {Autoencoders on field-programmable gate arrays for real-time,
  unsupervised new physics detection at 40 {MHz} at the Large Hadron Collider}.
\newblock {\em Nature Machine Intelligence}, 4(2):154--161, 2022.

\bibitem{bucilua2006model}
C~Bucilua, R~Caruana, and A~Niculescu-Mizil.
\newblock Model compression.
\newblock In {\em proceedings of the 12 th ACM SIGKDD International Conference
  on Knowledge Discovery and Data Mining}, volume~3, 2006.

\bibitem{hinton2015distilling}
G~Hinton, O~Vinyals, and J~Dean.
\newblock Distilling the knowledge in a neural network, 2015.

\bibitem{stanton2021does}
S~Stanton, P~Izmailov, P~Kirichenko, AA~Alemi, and AG~Wilson.
\newblock Does knowledge distillation really work?, 2021.

\bibitem{salehi2021multiresolution}
M~Salehi, N~Sadjadi, S~Baselizadeh, MH~Rohban, and HR~Rabiee.
\newblock Multiresolution knowledge distillation for anomaly detection.
\newblock In {\em Proceedings of the IEEE/CVF conference on computer vision and
  pattern recognition}, pages 14902--14912, 2021.

\bibitem{liu2022unsupervised}
H~Liu, K~Li, X~Li, and Y~Zhang.
\newblock Unsupervised anomaly detection with self-training and knowledge
  distillation.
\newblock In {\em 2022 IEEE International Conference on Image Processing
  (ICIP)}, pages 2102--2106. IEEE, 2022.

\bibitem{xiao2021unsupervised}
Q~Xiao, J~Wang, Y~Lin, W~Gongsa, G~Hu, M~Li, and F~Wang.
\newblock Unsupervised anomaly detection with distillated teacher-student
  network ensemble.
\newblock {\em Entropy}, 23(2):201, 2021.

\bibitem{abnar2020transferring}
S~Abnar, M~Dehghani, and W~Zuidema.
\newblock Transferring inductive biases through knowledge distillation.
\newblock {\em arXiv preprint arXiv:2006.00555}, 2020.

\bibitem{emd}
LN~Vaserstein.
\newblock Markov processes over denumerable products of spaces, describing
  large systems of automata.
\newblock {\em Problemy Peredaci Informacii}, (3):64--72, 1969.

\bibitem{kulkarni2017knowledge}
M~Kulkarni, K~Patil, and S~Karande.
\newblock Knowledge distillation using unlabeled mismatched images.
\newblock {\em arXiv preprint arXiv:1703.07131}, 2017.

\bibitem{sau2016deep}
BB~Sau and VN~Balasubramanian.
\newblock Deep model compression: Distilling knowledge from noisy teachers.
\newblock {\em arXiv preprint arXiv:1610.09650}, 2016.

\bibitem{mirzadeh2020improved}
SI~Mirzadeh, M~Farajtabar, A~Li, N~Levine, A~Matsukawa, and H~Ghasemzadeh.
\newblock Improved knowledge distillation via teacher assistant.
\newblock In {\em Proceedings of the AAAI conference on artificial
  intelligence}, volume~34, pages 5191--5198, 2020.

\end{thebibliography}

\end{footnotesize}

% % ****************************************************************************
% % END OF BIBLIOGRAPHY AREA
% % ****************************************************************************

\end{document}